# A CONVOLUTIONAL NEURAL NETWORK-BASED PATENT IMAGE RETRIEVAL METHOD FOR DESIGN IDEATION

Shuo Jiang[12*], Jianxi Luo[3], Guillermo Ruiz Pava[2], Jie Hu[1], Christopher L. Magee[2]

[1]Shanghai Jiao Tong University, 800 Dongchuan Rd, Shanghai 200240, China
[2] Massachusetts Institute of Technology, 77 Massachusetts Avenue, Cambridge MA 02139, USA
[3] Singapore University of Technology and Design, 8 Somapah Rd 487372, Singapore

**ABSTRACT**

*The patent database is often used in searches of inspirational stimuli for innovative design opportunities because of its large size, extensive variety and rich design information in patent documents. However, most patent mining research only focuses on textual information and ignores visual information. Herein, we propose a convolutional neural network (CNN)-based patent image retrieval method. The core of this approach is a novel neural network architecture named Dual-VGG that is aimed to accomplish two tasks: visual material type prediction and international patent classification (IPC) class label prediction. In turn, the trained neural network provides the deep features in the image embedding vectors that can be utilized for patent image retrieval and visual mapping. The accuracy of both training tasks and patent image embedding space are evaluated to show the performance of our model. This approach is also illustrated in a case study of robot arm design retrieval. Compared to traditional keyword-based searching and Google image searching, the proposed method discovers more useful visual information for engineering design.*

## 1. INTRODUCTION

Engineering designers often reuse and transform previous technical knowledge and ideas into their new designs [1–3]. Having a broad understanding of the precedents is essential to exploring the full design space and improving design ideation. Among different kinds of data sources of design precedents, the patent repository is probably the biggest digitalized design database. In a patent, graphical information plays an important role to present its novelty. However, in comparison with text information, the current patent retrieval methods or systems disregards the image information. Furthermore, some innovative design features are only presented in patent drawings but not shown in the text. New technologies and innovations are often described with rapidly changing and inconsistent terms, but drawings could present similar characteristics [4]. Design ideation is to generate as many ideas as possible, it is beneficial to inspire designers directly and effectively at the early stage of the design process. Patent image retrieval techniques have therefore become crucial to engineering design ideation.

Some prior works have mined and analyzed patent images [5–9]. Although it can provide many potential benefits for data-driven design, patent image retrieval faces some challenges. First of all, the patent database is vast. For example, the United States patent and trademark office's (USPTO) patent database alone contains more than ten million patent records. A patent has more than ten images in average. Identifying relevant images to the given image in the database is not a trivial task. Second, all the images found in patent files are binary (black and white). Furthermore, the knowledge within patent images is hard to represent due to their texture-less characteristics. The current image searching methods are mostly based on content-based image retrieval (CBIR) techniques, which exploit the color, shape and textural information of images [10,11]. Some researchers use local features, such as scale-invariant feature transform (SIFT) features, to represent a patent image [4], but the retrieval results only utilize geometric information of the query image, ignoring domain knowledge. Compared to the classical local descriptors, CNN enables learning richer features of images because of its deep architecture [12]. Based on the more informative features extracted from CNN, retrieval method would return more relevant results to the given image.

In this paper, we introduce a CNN-based approach for patent image retrieval for engineering design ideation. The goal of this research is to retrieve patent images, which are not only visually similar to the given query but also share similar design knowledge with the given query. We propose a novel convolutional neural network architecture named Dual-VGG. It is trained with two different tasks: visual material type prediction and IPC class label prediction. Specifically, visual material type prediction is to classify patent images into 9 pre-defined type categories [6], namely abstract drawing, flowchart, graph, chemical structure, table, DNA, math formula, computer program and symbol. Thus, the first task enables our model to learn visual characteristics of images. In the IPC, the technical topics are categorized based on the structure and function of the artifacts and supplemented by their applications. In this way, the IPC labels are not only the index of different groups. They contain domain knowledge of structures, functionality, effects and principles that might be useful for the design inspiration process [13]. Thus, the second task, the IPC class label prediction, enables our model to learn domain knowledge. After the training and validating process, the deep features can be extracted from the last second-to-last fully connected layer of the neural network for the further utilization.

In the design-by-analogy field, the relationship between design ideation effects and the distance of inspirational stimuli is also well studied [14]. Goucher-Lambert et al. [15] revealed that the stimuli that is close to the initial design statement (near stimuli) improved the usefulness and feasibility of design

---

solutions, whereas far stimuli improved the novelty. Goucher-Lambert et al. [16] and Chan et al. [17] separately pointed out that near stimuli may generally lead to better design outcomes than far stimuli. Thus, in this research, the retrieved patent images can be directly viewed as the near stimuli for designers. Based on the features extracted from the trained network, we then use the cosine similarity method to compute pairwise similarities between a given query image and other images to support the search and retrieval of most relevant images to the query.

This research contributes to the growing studies about data-driven design [1,18,19], design-by-analogy [20,21] and machine learning-based design analytics [22–26]. The main contributions of this research are twofold: 1) A novel convolutional neural network architecture, which are trained to predict both visual material type and IPC class label of patent images, which forms a patent image embedding space and provides vector representations of any given images; 2) A patent image retrieval approach using the trained model, which can be used for engineering design aid and ideation based on the patent database.

The paper is organized as follows. Section 2 reviews the literature on data-driven design, convolutional neural networks and image retrieval. In Section 3, we present Dual-VGG convolutional neural network, deep feature extraction and patent image retrieval methodology. In Section 4, we describe the experiment, including the dataset, model training process, experimental results and discussion. Then we present the application cases of the proposed method in Section 5. Finally, we conclude with discussions on limitations of our methodology and discuss future work in Section 6.

## 2. LITERATURE REVIEW

This paper builds on prior works on patent data mining to develop a CNN-based patent image retrieval method to support design ideation and exploration. Herein we review relevant literature about data-driven design, and patent image retrieval.

### 2.1 Data-driven design ideation

Several studies have been done in the engineering design field to aid designers for design ideation and exploration based on big data techniques. For example, design-by-analogy aims to extract functional analogies from general data sources to help designers in systematically searching and identifying relevant cases or knowledge [20,21]. The theory of inventive problem solving (TRIZ) is a problem-solving and analysis tool developed from the study of patterns in the global patent documents [27]. The infused design method uses the metalevel representation of design problems to facilitate the discovery and use of knowledge across different technological fields [28]. Reich and Shai [29] proposed the engineering knowledge genome to enhance the searching of common knowledge and method structures in interdisciplinary domains. A few data-driven computer tools have been developed to implement these design theories. Amaresh et al. [30] created the Idea-Inspire system for designers to retrieve descriptions of systems from artificial and natural fields that could act as stimuli for design ideation to solve the problem. Goel et al. [31] developed the design by analogy to nature engine (DANE) tool to identify specific kinds of analogical design activity, as well as to facilitate study into the cognitive foundations of analogical design. AskNature is a web-based tool [32], with more than 1,600 biological strategy cases in its repository, to provide biological inspirations to design.

Specifically, as the largest digitalized design repository, the patent database is widely used to develop design methods using various data mining techniques. For example, Cascini et al. [33] developed a system named Pat-Analyzer to automatically identify the design contradictions behind a patented invention in TRIZ by analyzing patent textual information. Mukherjea et al. [34] created a biomedical patent semantic web that can recommend patents based on the semantic relationship between biological terms from the patent text. Ji et al. [13] proposed a design inspiration method which uses the knowledge from international patent classification of US patents for design education. Luo et al. [19,35] developed a computer-aided ideation system called InnoGPS using a network map of all patent classes to augment the exploration of design opportunities in the total technology space and to guide the retrieval of prior knowledge and concepts across domains for design analogies and syntheses. Sarica et al. [22] developed a semantic network of more than 4 million technical terms named TechNet based on the USPTO database for engineering knowledge discovery and ideation support. However, these prior methods are aimed to mine bibliographic and classification information of patents as well as patent textual data, ignoring image information. We should notice that many technological specifications are more appropriately described by technical figures in many domains, especially mechanical design. Thus, retrieving design information in images of a patent document can play a crucial role and enhance patent analytics for data-driven design.

### 2.2 Patent image retrieval

A few researchers have proposed patent image retrieval methods, and most of them are SIFT-based. These approaches often refer to a 128-dimensional descriptor. With a pre-trained vocabulary list, local features are projected to visual words. Thus, we can represent images in a similar format to documents, and classic indexing and searching schemes can be leveraged [36]. Huet et al. [7] proposed a relational skeleton method to capture the geometric features of the patent image based on the relationship between captured line segments. However, this system works well for some images while poorly for others. And the database used in this system is quite small. Vrochidis et al. [6] created the adaptive hierarchical density histogram (AHDH) to exploit the content of patent images. This method has been extended by Sidiropoulous et al. [37], who introduced the quantized relative density information. Both studies calculate the distribution of the binary pixels in a patent image by performing image segmentation. Csurka et al. [5] used Fisher Vectors to represent patent images and then compute the pairwise similarity between the images based on their dot product.

In recent years, CNN-based methods have shown advantages over such SIFT-based models which exploit simple geometric features within patent [12]. Features extracted from CNN have been shown to outperform geometric features in many computer vision tasks, such as visual object localization, image segmentation and image classification. In the image retrieval field, some competitive results from the use of CNN compared to traditional SIFT-based features have already been reported, even with low dimensional vectors [38–41]. These CNN-based approaches usually extract compact representations from a specific fully connected layer in the CNN architecture and then employ the approximate nearest neighbor or Euclidean distance-based searching method for retrieval. However, these prior studies only focused on general images rather than technical images, such as patent images, which are usually colorless and intricate.

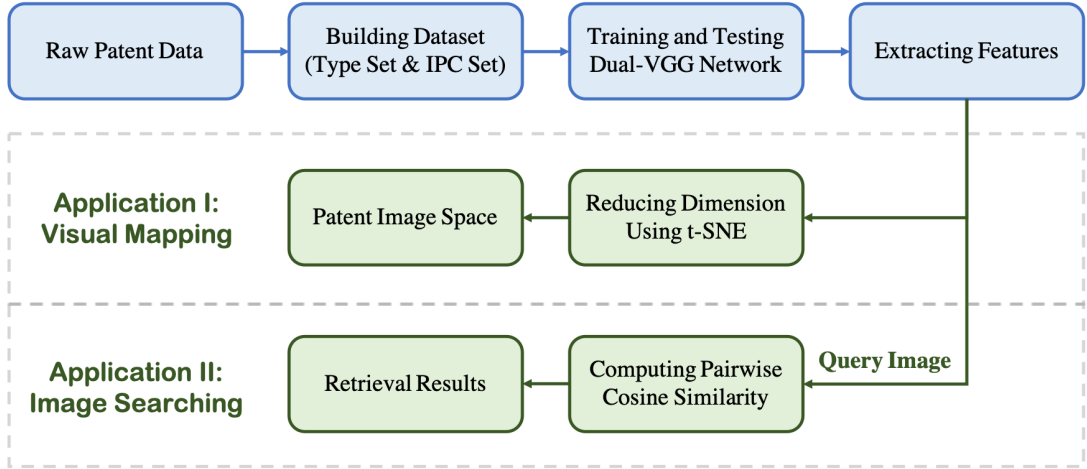

**Figure 1.** The overall research framework of the proposed method.

Taken together, prior research of CNN has made it possible for us to develop a new patent image searching method based on CNN features for data-driven engineering design. In the following chapter, we will present the methodology in detail.

## 3. METHODS

Our methodology includes training a novel convolutional neural network named Dual-VGG network to extract the deep features from patent images, and searching similar patent images using the cosine similarity method when given a query image related to a specific engineering design problem. Specifically, we first present the Dual-VGG network architecture. The network is aimed to predict the IPC class label for each patent, which equals to a classification task. In this study, we use the 1-digit IPC class label for each of the eigh patent classes from A to H, while each class comprises of smaller 3 to 7-digit subclasses in a hierarchy. Although one patent may belong to more than one class, we choose the bigger class as its label. The work can be viewed as a multi-class single-label classification task. After the training process, we can extract features from the second-to-last layer of the network, which is a fully connected layer. Then we can take all patent images in our database and represent each one as a specific vector. The dimension of vectors can be fixed on any given number, usually 512 or 1,024. In this work, we set 1,024 nodes for this fully connected layer. The cosine similarity method is used to calculate the pairwise similarity between two images. In this way, we can find the nearest neighbors in the embedding space, which are also the most similar patent images to a given image.

The whole framework of this research is shown in Figure 1. The Dual-VGG network lies at the heart of the method. It enables learning rich image embeddings, containing not only basic knowledge of shapes, colors and patterns but also domain knowledge in IPC information. The quality of features is mostly dependent on the performance of the convolutional neural network.

### 3.1 Dual-VGG convolutional neural network

The Dual-VGG network is designed based on the VGG19 convolutional neural network [42], which is proposed by Visual Geometry Group of the University of Oxford. The VGG19 is a robust convolutional neural network, which is widely used in many applications related to computer vision. The VGG19 has 19 trainable layers, including convolutional layers, fully connected layers, max-pooling layers as well as dropout layers. In the result section, we will evaluate the performances of VGG19 against alternative network architectures.

The Dual-VGG includes two VGG19 blocks, shown in Figure 2. The lower one is for visual material type classification. Specifically, we add an auxiliary task to classify each patent image into a specific type group, which can also be viewed as a multi-class classification task. According to literature [6], there are nine types for patent images, including drawing, flowchart, graph, chemical structure, table, DNA, math formula, computer program and symbol. During the training phase, we first train the lower VGG network independently from the pre-trained weights. Then we freeze all the weights in the lower VGG. The output vector from this block will contain information related to visual characteristics. Meanwhile, the upper branch of the network aims to learn both general geometrical knowledge and deep knowledge related to IPC information.

In the network, we resize the patent image into a $224 \times 224 \times 3$ vector for the input layer. There are several convolutional layers, activation layers as well as max-pooling layers in both VGG19 blocks. Convolutional layers represent a convolution operation over the image using a feature map and perform it at each point, then pass the result to the next layer [43].

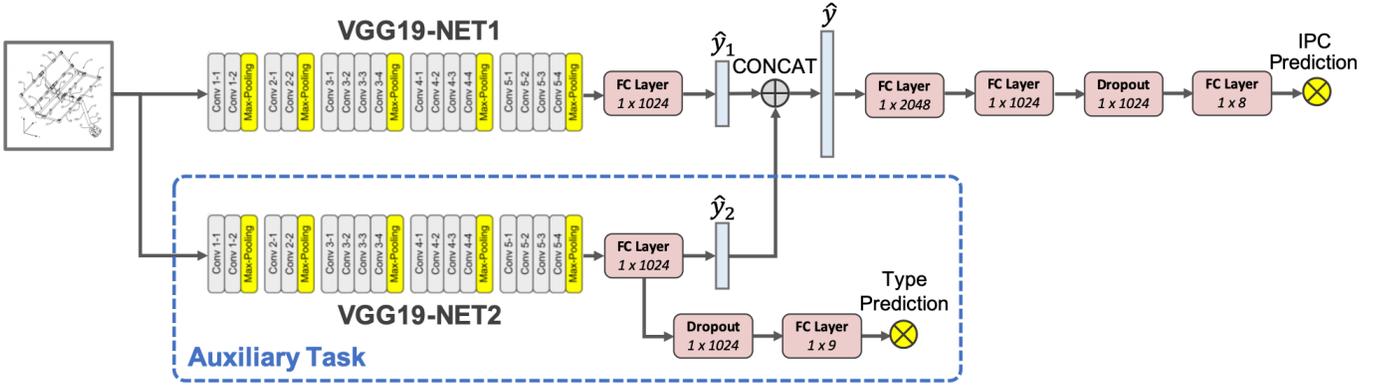

**Figure 2.** The Dual-VGG network architecture.

Filters belonging to the convolutional layers are some trainable feature extractors of usually a $3 \times 3$ size. Each convolutional layer stack would be followed by an activation function and max-pooling operation. The activation function used here is rectified linear unit (ReLU), which is a computationally efficient nonlinear function. After the activation function, a max-pooling layer is applied to our model. We take a filter of $2 \times 2$ size and a stride of the same length. And the output value is the max number in every region that the filter convolves around [44]. In the fully connected layer, each neuron receives an input from all the neurons in the last layer. A fully connected layer provides learning features from all the information of the previous layer, whereas a convolutional layer is usually based on consistent features with a small repetitive region. Through two VGG blocks, two 1,024-dimensional feature vectors are obtained. The concatenation layer connects two vectors into a 2,048-dimensional vector. Finally, after the concatenation layer, another fully connected layer takes all features in the previous layer and generates the probabilities distribution for each IPC class label.

Formally, in our model, let the patent image be $I_i$. We trained a specialized Dual-VGG network $f_v = I_i \rightarrow u_i$. For every patent image $I_i$, $i$=1 to $N$, we can use our model to extract its features as a vector $u_i$. In this way, we converted the image space $I$ into a latent feature space $R_d$. During the training process, the loss is calculated based on the categorical cross-entropy, and the back-propagation algorithm is used to change the weights and biases to make the network more predictive. The objective function is:

$$L = -\frac{1}{N}\sum_{i=1}^{N}\sum_{j=1}^{K}(y_j^i \log(\hat{y}_j^i))$$

where $N$ represents the number of images in a batch. $i$ means the index of an image. $K$ means the number of classes, in this case, we have $K = 8$. $y_j$ represents the output value of the last fully connected layer and we have $\sum_{j=1}^{K} y_j = 1$. $\hat{y}_j$ means the ground truth; it equals to 1 for the true class and 0 for others.

After the training process, the model can be used for both classification and feature extraction purposes. Since the goal of the model is to classify patent images into a specific IPC 1-digit class, the features extracted from the network will contain technical domain knowledge learned from IPC information.

### 3.2 Deep feature extraction and patent image retrieval

Using the Dual-VGG network, we can transform the patent images into feature vectors. Specifically, each image is converted into a 1,024-dimensional vector from the second-to-last fully connected layer of the network. It is noteworthy that the number of nodes in that layer can be modified as required. For example, we can also set 512 nodes and then obtain 512-dimensional vectors. Besides, our Dual-VGG network is capable of learning input data of different sizes, which means our learning model can be updated easily over time.

Based on the feature representations, we can compute the pairwise similarity between two feature vectors $u_i$ and $u_j$ using the cosine similarity:

$$sim(u_i, u_j) = \frac{u_i^T u_j}{\|u_i\|\|u_j\|}$$

Here, value 1 is when patent images are identical, and 0 (in the case of non-negative 1,024-dimensional feature vectors) or below 0 when the images are completely dissimilar [45].

The vector space and the associations of vectors enable at least two strategies of patent image retrieval for design inspiration. First, given a query patent image, we can compute its pairwise similarities to all other images and identify the most similar images to it. This represents a direct image-to-image search. Second, given a language-based design query, we can first use keywords to query for relevant patents using the USPTO website, Google Patents, Patsnap, Incopat, or other patent search tools, and then identify other images in our trained patent image database that are most similar to the figures in the retrieved patent set. Both strategies can be used to discover visual design knowledge for inspiration, in the conceptual design ideation and prior art searches. Now the design knowledge retrieval and discovery are conducted through the inference of design knowledge embedded in images.

### 4. EXPERIMENTS
### 4.1 Data collection and preprocessing

There are three parts of the data in this work: a dataset about image types used in the auxiliary task of the Dual-VGG network, an IPC class label dataset used in the main task and an evaluation dataset including thousands of patent images from four technical domains.

The dataset used in the auxiliary task comes from the CLEF-IP 2011 [46], including more than 40,000 images. All the images come from European Patent Office (EPO) patents and three IPC classes that have an application date before 2002. The original dataset contains from 300 to 6,000 training images for each of nine categories, which means it is an unbalanced dataset. In order to improve the final performance of the classifier, a data augmentation technique was used to enrich our training data. Data augmentation methods such as random rotations or nonlinear deformations are effortless to implement and effective at improving classification accuracy in some settings [47]. After data augmentation, a balanced and enlarged dataset was obtained compared to the original CLEF-IP 2011 dataset. The training set, validation set, and test set were separated as 6:1:1. Thus we have respectively 6,000 images, 1,000 images, 1,000 images in training, validation and test processes for each of nine classes.

The IPC class label dataset is made up of all USPTO granted patent database from 2005 to 2009. The raw patent data were downloaded from the online website of the USPTO[1]. The bulk data includes the full texts of patents since 1976 and figures of patents in TIFF format since 2004. We collected 403,870 utility patents in 8 categories at the IPC 1-digit level and their 10,877,766 images. It is noteworthy more than one figure may be shown on the same page. In this case, we used a fine-tuned YOLO-V3 neural network [48] to detect figures and segment them. Then, 13,998,254 figures were extracted in total. We gave the same IPC label for different figures of the same patent.

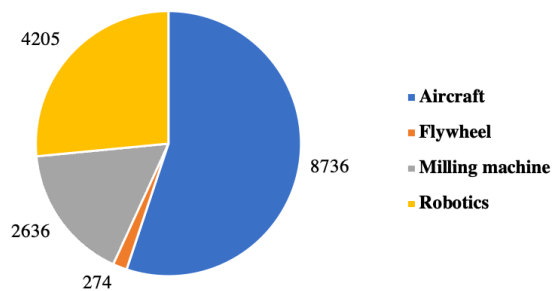

**Figure 3.** Domain distribution of patents in the dataset

To evaluate the performance of the representation learning and carry out an application example, we created an evaluation dataset containing patents in four different technological domains. The classification overlap method (COM) developed by Benson and Magee [49,50] was used to obtain a relatively relevant and complete patent set of a certain specific domain. COM had been successfully applied in several prior studies [51,52]. We selected four domains related to mechanical engineering for our dataset: flywheel, milling machine, aircraft transport and robotics. The patents in each domain set have an approximately 80% relevant rate, which is exactly suitable for our purpose of retrieval experiments to involve some irrelevant data. 15,851 patents were obtained from these four domain sets. Then we downloaded all the figures of these patents from the PATSNAP[2]. Only images on the front page were used in this dataset with the assumption that they are more important and representative for the whole patent than other figures. The distribution of patents in different domains is shown in Figure 3. This dataset was used to implement a case study on robot arm design retrieval and exploration.

**4.2 Neural network training**

The neural network architecture and classification process have been implemented in Python using Keras library[3], which provides an application interface for loading pre-trained models. We first trained a half part of the model aimed to the auxiliary task individually and froze all the weights after training. Then we trained the rest of the model. During the training process, the dataset performs forward pass and backward pass to update the parameters of the model, including both weights and bias. When forward and backward pass finish passing the whole dataset, it is called one epoch. For other hyperparameters, we set 1e-4 for the learning rate of the SGD optimizer and 32 for the batch size.

The network was trained for 50 iterations for the auxiliary task, and 10 iterations for the IPC class label prediction task. The training process for each task took 16.8 hours and 136.2 hours respectively. We employed the 4 GPUs (2 Titan Xp 16GB and 2 Titan N 16GB) with the Intel core i7 processor to train deep learning models more rapid.

**4.3 Results and discussion**

In this work, we trained the Dual-VGG neural network for classifying patent images into different visual material types (called Auxiliary task below) and explicit IPC 1-digit label classes (called Main task below). For a benchmarking and evaluation purpose, we also implemented the same datasets and tests, using ResNet50 [53], DenseNet121 [54] as well as XRCE's method [5].

ResNet50 and DenseNet121 are the most popular and successful convolutional neural networks for image classification tasks. XRCE's method is the 1st place winner of the image type classification task in the CLEF-IP 2011 contest which achieved 92.2% accuracy. XRCE's method represented images with Fisher Vectors and trained many one-versus-all linear classifiers. Table 1 shows the accuracy of the test data of each model. Dual-VGG outperforms all other models in both auxiliary and main tasks. Three CNN-based methods are better than XRCE's method in the type classification task. This is because convolutional neural networks can learn richer features of images due to its deep architecture compared to the classical local descriptors. Thus, CNN-based retrieval methods have been continuously proposed in recent years and are gradually replacing the handcrafted local detectors and descriptors [12]. As for the main task, Dual-VGG performs better than the other two neural networks, showing the highest accuracy of 54.32%. This is because the Dual-VGG network combines the auxiliary task and main task in the same model architecture, while ResNet50 and DenseNet121 implement two tasks independently. It shows that involving type information into the model is useful to predict the IPC 1-digit classes. However, we should note that the accuracy (54.32%) is not that high. We believe there are two reasons. One reason is that it is naturally difficult to predict the IPC class label only with single patent images even for human experts. For example, in some figures, only the partial view of a specific design is shown, which cannot represent the full design information of the patent. Another is that the resolutions of some patent images are low.

---

[1] https://bulkdata.uspto.gov
[2] https://www.patsnap.com
[3] https://github.com/keras-team/keras

**Table 1.** Classification accuracy of Dual-VGG, ResNet50, DenseNet121 and XRCE's method on the test set. The auxiliary task is to predict the visual material type class and the main task is to predict the IPC class label of patent images.

| Method | Accuracy | |
|---|---|---|
| | Auxiliary task | Main task |
| Dual-VGG | **95.37%** | **54.32%** |
| ResNet50 | 92.54% | 49.67% |
| DenseNet121 | 94.69% | 52.45% |
| XRCE's | 92.20% | / |

Figure 4 shows the confusion matrix of the Dual-VGG network. In Figure (a), it is clear that our model performs well for nearly all visual material types except tables. Figure (b) shows our model learned more information from IPC classes A, B, C, F, G and H compared to D and E. This is because that the dataset used for training is unbalanced. We can observe that the model ignores class D as this group is quite small. The two confusion matrices show that our model learned specific knowledge about type and IPC information, which is useful for tasks of relating and mapping images as well as image search.

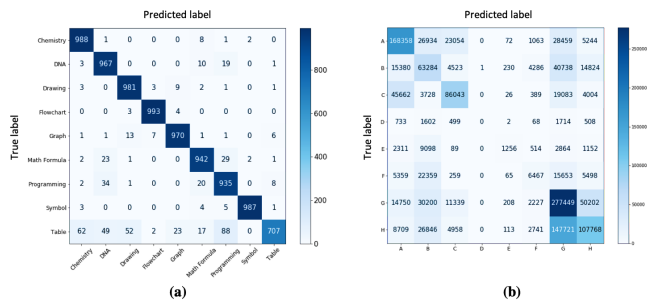

**Figure 4.** (a) Confusion matrix of the auxiliary task; (b) Confusion matrix of the main task.

## 5. APPLICATION CASES
### 5.1 Visual mapping of patent image space
Figure 5 shows the patent image space generated by our model. The t-SNE [55] algorithm is used to project 1024-dimensional image features into a 2D space. Since the convolutional neural network tends to match visually similar images to nearby places in the latent space, we can see that the images of similar design space (e.g., flywheels) are projected close to one another. However, even when patent images are not visually similar, they might be shown together if they shared identical knowledge from the IPC information.

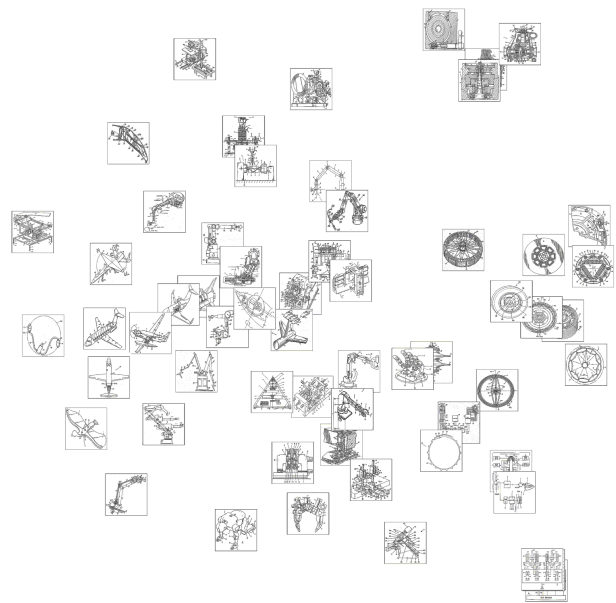

**Figure 5.** Visualization of patent image space in 2D space using t-SNE.

We can observe that in the left part of the figure, various forms of aircraft images gather together, while many flywheel images are clustered in the right section. Also, there are some flowcharts shown in the nearby place. Remarkably, there is an image with circle-shape shown near the airplane in the left part of the space. This image comes from a patent named Ejection System, describing a system to impart the desired combination of linear and rotational momentum to a projectile. Although its shape is similar to a flywheel, it is not projected to the place near those flywheels because domain knowledge has already been embedded in its feature vector. This visual map gives us a glimpse of the neural network model and in part justifies that our model can project the patent images into a reasonable embedding space.

### 5.2 Robot arm design retrieval and ideation
Nowadays, robots are increasingly used in various working environments to replace manual work, especially to perform repetitive tasks. The robot arm is a kind of robot manipulator, usually programmable, with similar functions to the real human arm. The manipulators' links are usually connected by joints, which allow rotational and translational motion. Those links can be regarded as a kinematic chain. The end of a kinematic chain in the manipulator is called the end effectors and is similar to human hands. The end effectors are usually designed to perform kinds of tasks depending on the application with high accuracy, such as welding, grasping and spinning. The robot arms can be manipulated manually or controlled automatically for industrial or home applications [56]. The history of robot arms dates back to 1959 in the US when George Devol designed the first robotic arm named 'Unimate' to lift and stack hot metal parts [57]. Since then, many industrial companies and academic institutions have made attempts to develop robot arms, such as KUKA, ABB, Hitachi, Boston dynamics, and so on.

Figure 6 shows the ten images most related to the keyword "robot arm" retrieved via a keyword search in Google Images. We can observe that these Google-retrieved images represent merchandise information from online shopping platforms or robot companies, such as Amazon, Ufactory and Ozrobotics. The appearance and functions of these robot arms are highly identical. Such retrievals do not provide broad design concepts nor sufficient engineering design information about robot arms to stimulate design ideation.

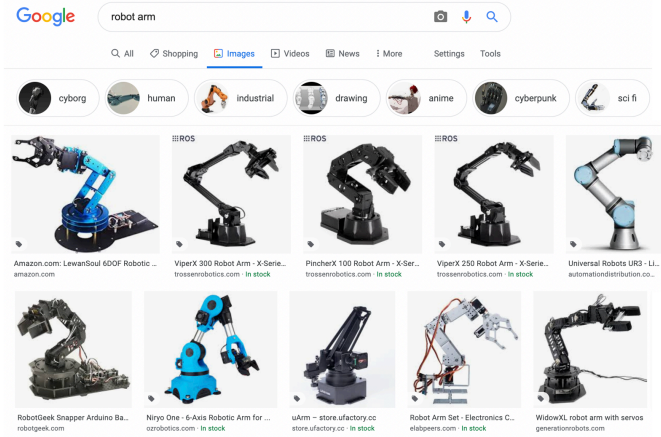

**Figure 6.** Most relevant images to "robot arm" as retrieved from Google Image.

For comparison, we conducted a keyword-based search in the USPTO patent database using PATSNAP and extracted patent images from these patents. Table 2 lists the top 10 relevant patents. Figure 7 shows the top 10 images extracted from the first page of these patents. It is evident that these returned patents are highly relevant as most of their titles include the query keyword "robot arm". However, these patent images in Figure 7 provide little visual design inspiration and designers would have to read the full texts of the patents that contain them. This is tedious and time-costly process. By contrast, it would be desirable if the retrieved images can directly inspire engineers with their visual information.

**Table 2.** Most relevant patents to "robot arm" as retrieved from Patsnap using keyword searching method.

| | Patent ID | Title |
|---|---|---|
| 1 | US10005182 | Robot arm |
| 2 | US4920500 | Method and robot installation for programmed control of a working tool |
| 3 | US4636138 | Industrial robot |
| 4 | US6430472 | Robot feature tracking devices and methods |
| 5 | US8290621 | Control apparatus and control method for robot arm… |
| 6 | US8185243 | Robot, control device for robot arm and control program for robot arm |
| 7 | US8909374 | Robot arm control apparatus, robot arm control method, robot... |
| 8 | US7040852 | Robot arm mechanism and robot apparatus |
| 9 | US10471596 | Robot arm and unmanned aerial vehicle equipped with the robot arm |
| 10 | US8340820 | Robot arm and method of controlling robot arm to avoid collisions |

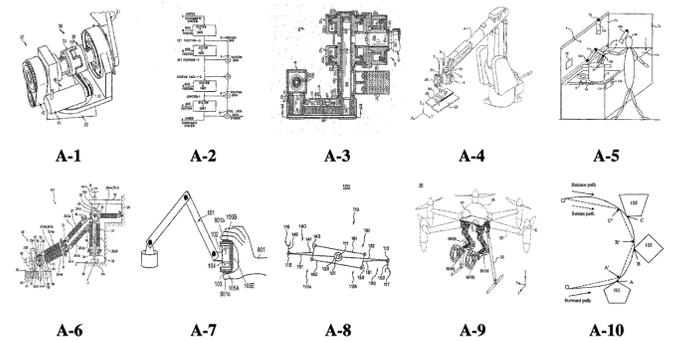

**Figure 7.** The patent images on the first page of the patents shown in Table 2.

Based on the proposed method in this paper and the trained Dual-VGG network, we have derived the 1,024-dimensional vectors for all the patent images in the database described in Chapter 4.1. We asked an engineer to select an initial query image manually, shown as Image B-0 in Figure 8, which is a coupler for connecting a robot arm of a remote manipulator device to a tool body and appears quite different from traditional robot arms. Then we identified its most similar images in the image embedding space according to pairwise similarities. Figure 8 shows the top 9 most relevant images obtained directly via the visual search, in contrast to those images in Figure 7 from the patent documents retrieved via a keyword search. Based on these images, we further identified their patent ID and titles, as listed in Table 3.

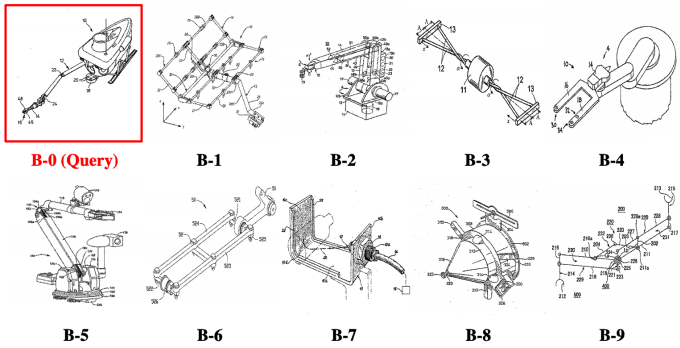

**Figure 8.** Top 9 most relevant patent images of the given query image (shown in the red box) from the database as retrieved by proposed method.

In Figure 8, Images B-2, B-4 and B-5 are in the traditional form of the robot arm, which comprise motors to drive the wrist mechanisms. Image B-1 presents a novel end-effector with the high flexibility, which is made up of several articulated booms. Image B-3 shows a motion conversion device used in robotic joints of robot arms. Image B-6 presents a moveable linkage for a parallel mechanism used in the mechanism design process of robot arms. Likewise, image B-9 is also a robot arm mechanism. It is remarkable that our method returned an image about pneumatic gripper, shown in Image B-7, which means patents comprising different functions can be retrieved. Image B-8 shows a grapple fixture with improved docking and engaging capabilities as an irregular design.

Patent images in Figure 8 show that the proposed CNN-based model enables us to obtain more visually sensible design information about robot arms with broader design concepts (functions, components, configurations, working principles, etc.) compared to those in Figure 7. These advantages are particularly valuable for augmenting design ideation. Although some of the returned patents do not include 'robot arm' term in their titles, as shown in Table 3, nearly all of them are actually relevant to the given topic. These patents in Table 3 are identified via a visual search, in contrast to those patents in Table 2 identified via a text search.

Moreover, we used another way to implement the patent image retrieval for robot arm design ideation. As described above, we obtained ten relevant patents and their images using a keyword search, shown in Figure 7. Then we identified four visually valuable items manually as the query images. Based on the proposed method, Figure 9 shows the retrieved relevant images in our database, which are most similar to the figures in the retrieved patent set. Compared to those in Figure 7, these images show more visually diverse and informative design information. With the keyword-to-image retrieval method and single query image-driven retrieval method, the design knowledge discovery and design opportunity exploration are augmented through the inference of design knowledge contained in related images.

**Table 3.** The patent ID and titles of the patents corresponding to the patent images in Figure 8.

| | Patent ID | Title |
|---|---|---|
| **Query** | US4227853 | Manipulator wrist tool interface |
| 1 | US8496425 | Reconfigurable end-effectors with articulating frame… |
| 2 | US4812104 | Electrical robot |
| 3 | US8256310 | Motion conversion device |
| 4 | US6942265 | Apparatus comprising a flexible vacuum seal pad structure… |
| 5 | US8322249 | Robot arm assembly |
| 6 | US8429998 | Parallel mechanism and moveable linkage thereof |
| 7 | US5066058 | Pneumatic gripper |
| 8 | US4929011 | Grapple fixture |
| 9 | US6705177 | Robot arm mechanism |

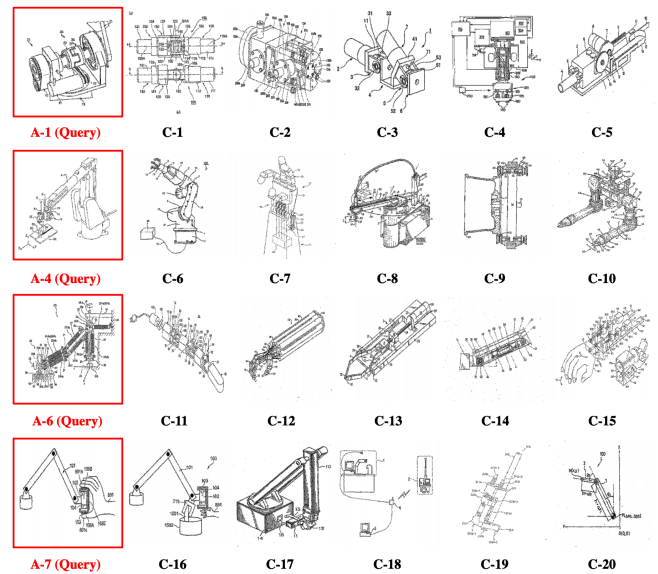

**Figure 9.** More relevant patent images of four query images in Figure 7 as retrieved by proposed method.

To evaluate the performance of the proposed method for design ideation, we conducted an evaluation experiment with 10 participants. All of them are second-year undergraduate students from the school of mechanical engineering of Shanghai Jiao Tong University. We assumed that all students were at similar levels of design creativity and experience. In the evaluation, we provided participants 39 patent figures from Figure 7 to Figure 9, which are retrieved by the simple keyword-based method and our proposed method. Then, all participants were asked to select 20 of them, which are relatively more informative and useful to the robot arm design task based in their evaluation. Figure 10 shows the visualization of the evaluation results. The red, green and blue markers represent patent images in Figure 7 (images A-1 to A-10), Figure 8 (images B-1 to B-9) and Figure 9 (images C-1 to C-20) respectively. In other words, they are obtained retrieval results based on different methods. From Figure 10, we can see that the green and blue markers have a higher density than the red ones, indicating the figures in groups B and C are more attractive to participants.

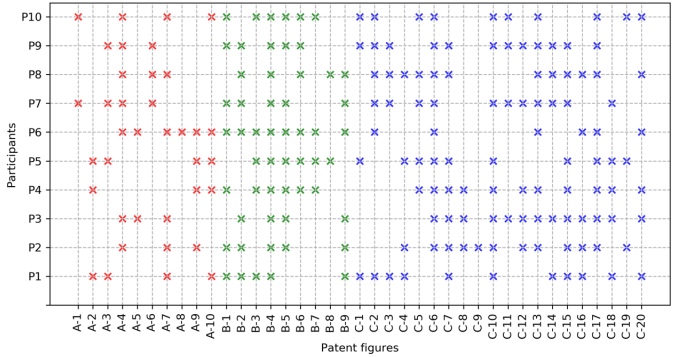

**Figure 10.** Visualization of the evaluation results.

To make the comparison clearer, we computed the overall evaluation score for three groups:

$$S_i = \frac{K_i}{N_i * M}, i = A, B, C$$

where $K$ denotes the number of marked figures in each group, $N$ is total figure count in each group, and $M$ is the number of evaluators. In this case, we have $N_A = 10$, $N_B = 9$, $N_C = 20$ and $M = 10$. A higher score indicates the higher probability that the participants would choose the figures from this group. This metric eliminates the influence of the group size. Evaluation scores are reported in Figure 11. The red dotted line represents the average score of the whole set. We can see that both scores of groups B and C are higher than the average score, while the score of group A is lower than it. The one-way analysis of variance (ANOVA) reports a p-value of 1.92E-11<0.05, which indicates the evaluation scores of different groups are statistically significant.

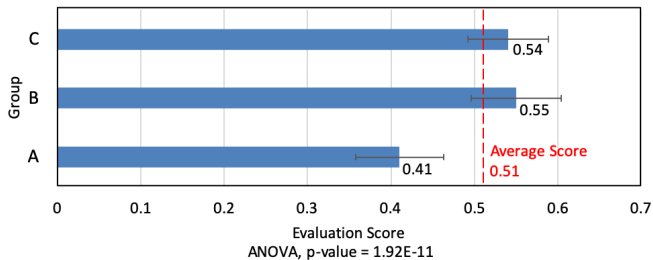

**Figure 11.** Evaluation scores of different image groups.

## 6. CONCLUDING REMARKS

In this paper, a CNN-based approach is developed to implement patent image retrieval for design ideation. The goal of this method is to retrieve patent images, which are not only visually similar to the given image but also share similar design knowledge with the given image. The proposed model enables learning both visual characteristics and domain knowledge through training at two different tasks: visual material type prediction and IPC class label prediction. With the trained model, two application cases are presented: the visual mapping of patent image space and robot arm design retrieval. This research contributes to the growing studies about data-driven design, analogy-based design and machine learning-based design analytics. In addition to the single case study shown in this paper, this method may also be useful for engineering education, technology information retrieval and innovation management.

This method still has some limitations in several aspects. First of all, we found that not every patent image can be used to represent the information of the whole patent. Some patent images are not in high quality. The retrieval results are related to the given query image to some extent. Second, in the visual map of images, we found that several images about robot arms were projected near the milling machines due to their similar shapes. Alternative network parameters in the training might potentially lead to better results. Furthermore, this paper presents only a single small-scale case study. To further validate the proposed method, more diverse contexts, more systematic tests, and more human evaluators are desirable.

In future work, on the one hand, we will build a more complete patent image database related to mechanical engineering domains and use the proposed method to derive a feature vector for every patent image for the wider uses of other interested researchers. On the other hand, we will continue to improve the neural network model and try to develop a more robust architecture to learn better features from patent images. Finally, we plan to combine both text information and visual information in patents to develop a more intelligent design knowledge retrieval system for design support.

## ACKNOWLEDGEMENTS

The authors acknowledge the funding support for this work received from the SUTD-MIT International Design Center and the China Scholarship Council (CSC). This research is also supported by Special Program for Innovation Method of the Ministry of Science and Technology, China (2018IM020100), National Natural Science Foundation of China (51975360, 51775332), National Social Science Foundation of China (17ZDA020). Any ideas, results and conclusions contained in this work are those of the authors, and do not reflect the views of the sponsors.

## REFERENCES

[1] Song, B., and Luo, J., 2017, "Mining Patent Precedents for Data-Driven Design: The Case of Spherical Rolling Robots," J. Mech. Des., **139**(11).

[2] Hatchuel, A., and Weil, B., 2009, "CK Design Theory: An Advanced Formulation," Res. Eng. Des., **19**(4), p. 181.

[3] Basnet, S., and Magee, C. L., 2016, "Modeling of Technological Performance Trends Using Design Theory," Des. Sci., **2**.

[4] Bhatti, N., Hanbury, A., and Stottinger, J., 2018,

"Contextual Local Primitives for Binary Patent Image Retrieval," Multimed. Tools Appl., **77**(7), pp. 9111–9151.

[5] Csurka, G., Renders, J.-M., and Jacquet, G., 2011, "XRCE's Participation at Patent Image Classification and Image-Based Patent Retrieval Tasks of the Clef-IP 2011.," *CLEF (Notebook Papers/Labs/Workshop)*.

[6] Vrochidis, S., Papadopoulos, S., Moumtzidou, A., Sidiropoulos, P., Pianta, E., and Kompatsiaris, I., 2010, "Towards Content-Based Patent Image Retrieval: A Framework Perspective," World Pat. Inf., **32**(2), pp. 94–106.

[7] Huet, B., Kern, N. J., Guarascio, G., and Merialdo, B., 2001, "Relational Skeletons for Retrieval in Patent Drawings," *Proceedings 2001 International Conference on Image Processing (Cat. No. 01CH37205)*, pp. 737–740.

[8] Tiwari, A., and Bansal, V., 2004, "PATSEEK: Content Based Image Retrieval System for Patent Database.," *ICEB*, pp. 1167–1171.

[9] Zhiyuan, Z., Juan, Z., and Bin, X., 2007, "An Outward-Appearance Patent-Image Retrieval Approach Based on the Contour-Description Matrix," *2007 Japan-China Joint Workshop on Frontier of Computer Science and Technology (FCST 2007)*, pp. 86–89.

[10] Datta, R., Joshi, D., Li, J., and Wang, J. Z., 2008, "Image Retrieval: Ideas, Influences, and Trends of the New Age," ACM Comput. Surv., **40**(2), pp. 1–60.

[11] Smeulders, A. W. M., Worring, M., Santini, S., Gupta, A., and Jain, R., 2000, "Content-Based Image Retrieval at the End of the Early Years," IEEE Trans. Pattern Anal. Mach. Intell., **22**(12), pp. 1349–1380.

[12] Zheng, L., Yang, Y., and Tian, Q., 2017, "SIFT Meets CNN: A Decade Survey of Instance Retrieval," IEEE Trans. Pattern Anal. Mach. Intell., **40**(5), pp. 1224–1244.

[13] Ji, Y., Qiu, Q., Feng, P., and Wu, J., 2019, "Empirical Study on the Impact of Knowledge in International Patent Classification on Design Inspiration of Undergraduate Students," Int. J. Technol. Des. Educ., **29**(4), pp. 803–820.

[14] Fu, K., Chan, J., Cagan, J., Kotovsky, K., Schunn, C., and Wood, K., 2013, "The Meaning of 'near' and 'Far': The Impact of Structuring Design Databases and the Effect of Distance of Analogy on Design Output," J. Mech. Des., **135**(2).

[15] Goucher-Lambert, K., and Cagan, J., 2019, "Crowdsourcing Inspiration: Using Crowd Generated Inspirational Stimuli to Support Designer Ideation," Des. Stud., **61**, pp. 1–29.

[16] Goucher-Lambert, K., Moss, J., and Cagan, J., 2019, "A Neuroimaging Investigation of Design Ideation with and without Inspirational Stimuli—Understanding the Meaning of near and Far Stimuli," Des. Stud., **60**, pp. 1–38.

[17] Chan, J., Dow, S. P., and Schunn, C. D., 2018, "Do the Best Design Ideas (Really) Come from Conceptually Distant Sources of Inspiration?," *Engineering a Better Future*, Springer, pp. 111–139.

[18] Kim, H. H. M., Liu, Y., Wang, C. C. L., and Wang, Y., 2017, "Data-Driven Design (D3)," J. Mech. Des., **139**(11).

[19] Luo, J., Yan, B., and Wood, K., 2017, "InnoGPS for Data-Driven Exploration of Design Opportunities and Directions: The Case of Google Driverless Car Project," J. Mech. Des., **139**(11).

[20] Murphy, J., Fu, K., Otto, K., Yang, M., Jensen, D., and Wood, K., 2014, "Function Based Design-by-Analogy: A Functional Vector Approach to Analogical Search," J. Mech. Des., **136**(10).

[21] Fu, K., Murphy, J., Yang, M., Otto, K., Jensen, D., and Wood, K., 2015, "Design-by-Analogy: Experimental Evaluation of a Functional Analogy Search Methodology for Concept Generation Improvement," Res. Eng. Des., **26**(1), pp. 77–95.

[22] Sarica, S., Luo, J., and Wood, K. L., 2020, "TechNet: Technology Semantic Network Based on Patent Data," Expert Syst. Appl., **142**, p. 112995.

[23] Rahman, M. H., Xie, C., and Sha, Z., 2019, "A Deep Learning Based Approach to Predict Sequential Design Decisions," *International Design Engineering Technical Conferences and Computers and Information in Engineering Conference*, p. V001T02A029.

[24] Hu, J., Qi, J., and Peng, Y., 2015, "New CBR Adaptation Method Combining with Problem--Solution Relational Analysis for Mechanical Design," Comput. Ind., **66**, pp. 41–51.

[25] Ma, J., Jiang, S., Hu, J., Shen, J., Qi, J., Tian, P., and others, 2018, "EXPLORING THE USE OF LONG SHORT-TERM MEMORY (LSTM) IN FUNCTIONAL BASED BIOINSPIRED DESIGN," *DS 89: Proceedings of The Fifth International Conference on Design Creativity (ICDC 2018), University of Bath, Bath, UK*, pp. 338–345.

[26] Panchal, J. H., Fuge, M., Liu, Y., Missoum, S., and Tucker, C., 2019, "Machine Learning for Engineering Design," J. Mech. Des., **141**(11).

[27] Al'shuller, G. S., and Shapiro, R. B., 1956, "On the Psychology of Inventive Creativity," Vopr. Psikhol, (6), pp. 37–49.

[28] Shai, O., and Reich, Y., 2004, "Infused Design. i. Theory," Res. Eng. Des., **15**(2), pp. 93–107.

[29] Reich, Y., and Shai, O., 2012, "The Interdisciplinary Engineering Knowledge Genome," Res. Eng. Des., **23**(3), pp. 251–264.

[30] Chakrabarti, A., Siddharth, L., Dinakar, M., Panda, M., Palegar, N., and Keshwani, S., 2017, "Idea Inspire 3.0—A Tool for Analogical Design," *International Conference on Research into Design*, pp. 475–485.

[31] Goel, A. K., Vattam, S., Wiltgen, B., and Helms, M., 2012, "Cognitive, Collaborative, Conceptual and Creative—Four Characteristics of the next Generation of Knowledge-Based CAD Systems: A Study in Biologically Inspired Design," Comput. Des., **44**(10), pp. 879–900.

[32] Deldin, J.-M., and Schuknecht, M., 2014, "The AskNature Database: Enabling Solutions in Biomimetic

Design," *Biologically Inspired Design*, Springer, pp. 17–27.

[33] Cascini, G., Russo, D., and others, 2007, "Computer-Aided Analysis of Patents and Search for TRIZ Contradictions," Int. J. Prod. Dev., **4**(1), pp. 52–67.

[34] Mukherjea, S., Bamba, B., and Kankar, P., 2005, "Information Retrieval and Knowledge Discovery Utilizing a Biomedical Patent Semantic Web," IEEE Trans. Knowl. Data Eng., **17**(8), pp. 1099–1110.

[35] Luo, J., Sarica, S., and Wood, K. L., 2019, "Computer-Aided Design Ideation Using InnoGPS," *International Design Engineering Technical Conferences and Computers and Information in Engineering Conference*, p. V02AT03A011.

[36] Lowe, D. G., 2004, "Distinctive Image Features from Scale-Invariant Keypoints," Int. J. Comput. Vis., **60**(2), pp. 91–110.

[37] Sidiropoulos, P., Vrochidis, S., and Kompatsiaris, I., 2010, "Adaptive Hierarchical Density Histogram for Complex Binary Image Retrieval," *2010 International Workshop on Content Based Multimedia Indexing (CBMI)*, pp. 1–6.

[38] Radenović, F., Tolias, G., and Chum, O., 2016, "CNN Image Retrieval Learns from BoW: Unsupervised Fine-Tuning with Hard Examples," *European Conference on Computer Vision*, pp. 3–20.

[39] Tolias, G., Sicre, R., and Jégou, H., 2015, "Particular Object Retrieval with Integral Max-Pooling of CNN Activations," arXiv Prepr. arXiv1511.05879.

[40] Kalantidis, Y., Mellina, C., and Osindero, S., 2016, "Cross-Dimensional Weighting for Aggregated Deep Convolutional Features," *European Conference on Computer Vision*, pp. 685–701.

[41] Gordo, A., Almazán, J., Revaud, J., and Larlus, D., 2016, "Deep Image Retrieval: Learning Global Representations for Image Search," *European Conference on Computer Vision*, pp. 241–257.

[42] Simonyan, K., and Zisserman, A., 2014, "Very Deep Convolutional Networks for Large-Scale Image Recognition," arXiv Prepr. arXiv1409.1556.

[43] Gatys, L. A., Ecker, A. S., and Bethge, M., 2016, "Image Style Transfer Using Convolutional Neural Networks," *Proceedings of the IEEE Conference on Computer Vision and Pattern Recognition*, pp. 2414–2423.

[44] Krizhevsky, A., Sutskever, I., and Hinton, G. E., 2012, "Imagenet Classification with Deep Convolutional Neural Networks," *Advances in Neural Information Processing Systems*, pp. 1097–1105.

[45] Baeza-Yates, R., Ribeiro-Neto, B., and others, 1999, *Modern Information Retrieval*, ACM press New York.

[46] Piroi, F., Lupu, M., Hanbury, A., and Zenz, V., 2011, "CLEF-IP 2011: Retrieval in the Intellectual Property Domain.," *CLEF (Notebook Papers/Labs/Workshop)*.

[47] Van Dyk, D. A., and Meng, X.-L., 2001, "The Art of Data Augmentation," J. Comput. Graph. Stat., **10**(1), pp. 1–50.

[48] Redmon, J., and Farhadi, A., 2018, "Yolov3: An Incremental Improvement," arXiv Prepr. arXiv1804.02767.

[49] Benson, C. L., and Magee, C. L., 2013, "A Hybrid Keyword and Patent Class Methodology for Selecting Relevant Sets of Patents for a Technological Field," Scientometrics, **96**(1), pp. 69–82.

[50] Benson, C. L., and Magee, C. L., 2015, "Technology Structural Implications from the Extension of a Patent Search Method," Scientometrics, **102**(3), pp. 1965–1985.

[51] Benson, C. L., Triulzi, G., and Magee, C. L., 2018, "Is There a Moore's Law for 3D Printing?," 3D Print. Addit. Manuf., **5**(1), pp. 53–62.

[52] Feng, S., and Magee, C. L., 2020, "Technological Development of Key Domains in Electric Vehicles: Improvement Rates, Technology Trajectories and Key Assignees," Appl. Energy, **260**, p. 114264.

[53] Zhang, X., Zou, J., He, K., and Sun, J., 2015, "Accelerating Very Deep Convolutional Networks for Classification and Detection," IEEE Trans. Pattern Anal. Mach. Intell., **38**(10), pp. 1943–1955.

[54] Huang, G., Liu, Z., Van Der Maaten, L., and Weinberger, K. Q., 2017, "Densely Connected Convolutional Networks," *Proceedings of the IEEE Conference on Computer Vision and Pattern Recognition*, pp. 4700–4708.

[55] Maaten, L. van der, and Hinton, G., 2008, "Visualizing Data Using T-SNE," J. Mach. Learn. Res., **9**(Nov), pp. 2579–2605.

[56] Gautam, R., Gedam, A., Zade, A., and Mahawadiwar, A., 2017, "Review on Development of Industrial Robotic Arm," Int. Res. J. Eng. Technol., **4**(03).

[57] Ajlouny, N. S., 1975, "Programmable Universal Transfer Device."